\documentclass[letterpaper]{article} % DO NOT CHANGE THIS
\usepackage{aaai2026}  % DO NOT CHANGE THIS
\usepackage{times}  % DO NOT CHANGE THIS
\usepackage{helvet}  % DO NOT CHANGE THIS
\usepackage{courier}  % DO NOT CHANGE THIS
\usepackage[hyphens]{url}  % DO NOT CHANGE THIS
\usepackage{graphicx} % DO NOT CHANGE THIS
\usepackage{amsmath}
\usepackage{caption}
\usepackage{amsfonts}
\usepackage{natbib}  % DO NOT CHANGE THIS AND DO NOT ADD ANY OPTIONS TO IT
\usepackage{caption} % DO NOT CHANGE THIS AND DO NOT ADD ANY OPTIONS TO IT
\usepackage{algorithm}
\usepackage{algorithmic}
\usepackage{subcaption}
\usepackage{xcolor}
\usepackage{booktabs}
\usepackage{array}
\usepackage{multirow}
\usepackage[pagebackref=true,breaklinks=true,letterpaper=true,colorlinks,bookmarks=false]{hyperref}     % hyperlinks
\usepackage{newfloat}
\usepackage{listings}
\DeclareCaptionStyle{ruled}{labelfont=normalfont,labelsep=colon,strut=off} % DO NOT CHANGE THIS
\floatstyle{ruled}
\newfloat{listing}{tb}{lst}{}
\floatname{listing}{Listing}
\title{SODiff: Semantic-Oriented Diffusion Model \\
for JPEG Compression Artifacts Removal}
\author{
    Tingyu Yang\equalcontrib\textsuperscript{\rm 1},
    Jue Gong\equalcontrib\textsuperscript{\rm 1},
    Jinpei Guo\textsuperscript{\rm 1,2},
    Wenbo Li\textsuperscript{\rm 3},
    Yong Guo\textsuperscript{\rm 4},
    Yulun Zhang\footnote{Corresponding author: Yulun Zhang, yulun100@gmail.com}\textsuperscript{\rm 1}
}
\affiliations{
    \textsuperscript{\rm 1}Shanghai Jiao Tong Univercity
    \textsuperscript{\rm 2}Carnegie Mellon University \\
    \textsuperscript{\rm 3}The Chinese Univercity of Hong Kong
    \textsuperscript{\rm 4}Max Planck Institute for Informatics
}

\usepackage{bibentry}
\begin{document}

\maketitle

\begin{abstract}
JPEG, as a widely used image compression standard, often introduces severe visual artifacts when achieving high compression ratios. Although existing deep learning-based restoration methods have made considerable progress, they often struggle to recover complex texture details, resulting in over-smoothed outputs. To overcome these limitations, we propose SODiff, a novel and efficient semantic-oriented one-step diffusion model for JPEG artifacts removal. Our core idea is that effective restoration hinges on providing semantic-oriented guidance to the pre-trained diffusion model, thereby fully leveraging its powerful generative prior. To this end, SODiff incorporates a semantic-aligned image prompt extractor (SAIPE). SAIPE extracts rich features from low-quality (LQ) images and projects them into an embedding space semantically aligned with that of the text encoder. Simultaneously, it preserves crucial information for faithful reconstruction. Furthermore, we propose a quality factor-aware time predictor that implicitly learns the compression quality factor (QF) of the LQ image and adaptively selects the optimal denoising start timestep for the diffusion process. Extensive experimental results show that our SODiff outperforms recent leading methods in both visual quality and quantitative metrics. Code is available at: \url{https://github.com/frakenation/SODiff}. 
\end{abstract}
\vspace{-3mm}
\vspace{-2mm}
\section{Introduction}
% \vspace{-1mm}
Joint Photographic Experts Group (JPEG)~\cite{Wallace1991JPEG} is a widely adopted lossy image compression standard that divides images into 8$\times$8 pixel blocks. JPEG then performs a discrete cosine transform (DCT) on each block, followed by quantizing high-frequency components to achieve compression. This method significantly reduces storage size while maintaining visually acceptable quality. However, under high compression, excessive quantization step sizes lead to over-quantization or even zeroing of DCT high-frequency coefficients. The loss of high-frequency information during reconstruction inevitably produces compression artifacts such as notable blocking and ringing. Previous deep learning-based methods~\cite{Dong2015ARCNN,Jiang2021FBCNN,liang2021swinir,zhang2018dmcnn} achieve considerable success in JPEG artifacts removal tasks. However, when facing such severe compression conditions, they often suffer from blocked details and distorted structures due to their lack of detail generation capability.

\begin{figure}[t]
\begin{center}
\includegraphics[width=0.85\columnwidth]{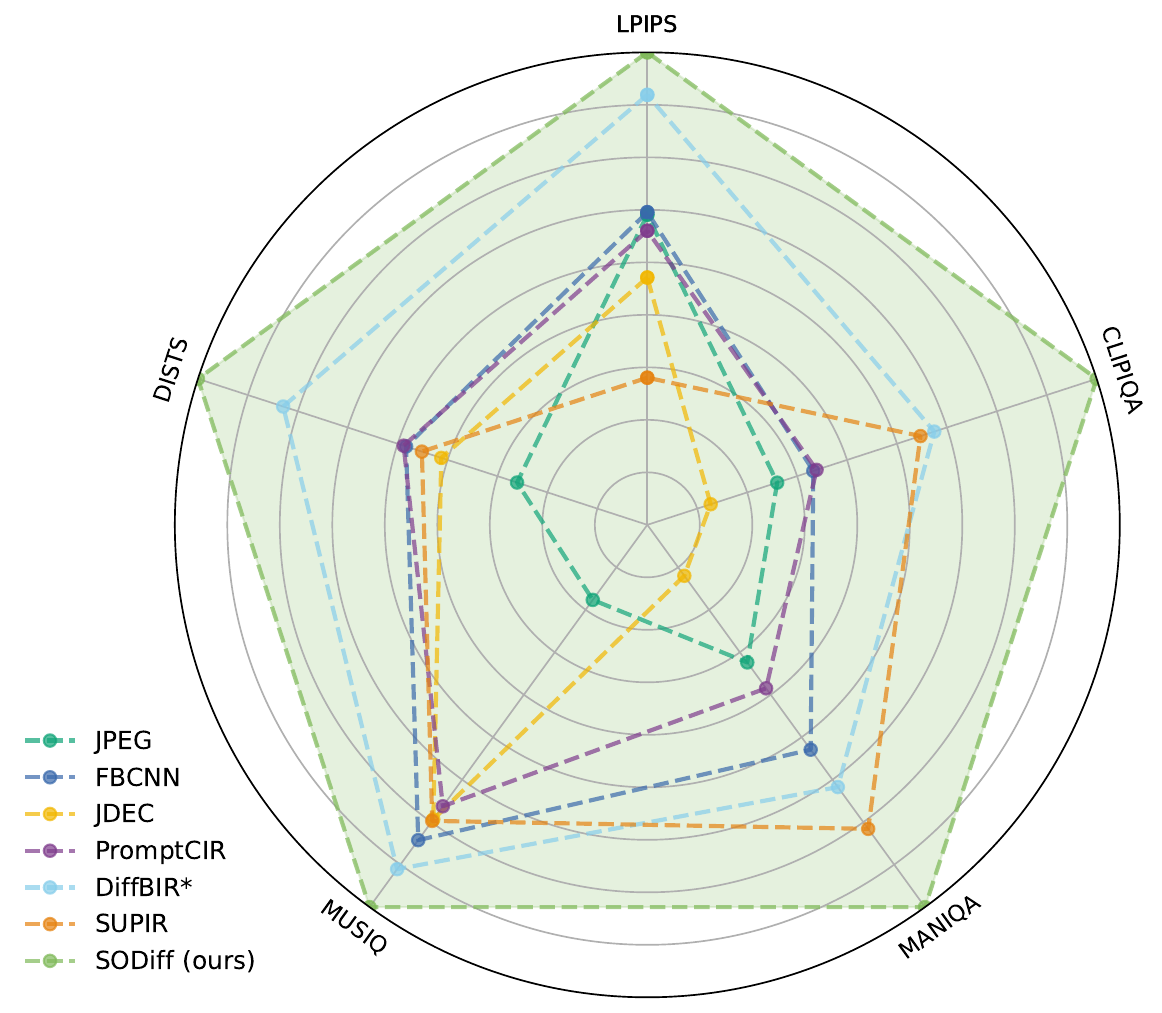}
\end{center}
\vspace{-5mm}
\caption{JEPG compression artifacts removal performance of recent methods on DIV2K-val (QF=5).}
\vspace{-8mm}
\label{fig:QF5_comparison_radar}
\end{figure}

In recent years, diffusion models (DM)~\cite{Rombach2022LDM,song2021ddim,Ho2020DDPM} based on latent space denoising mechanisms demonstrate excellent performance in image generation. Text-to-image (T2I) models~\cite{sd21,podell2023sdxl} generate high-quality images that match textual descriptions by encoding input text into semantic information to guide the denoising process. Pre-trained diffusion models possess powerful image generation prior knowledge. Related studies~\cite{lin2024diffbir,yu2024supir} show their superior performance in image restoration. Considering the time overhead of multi-step diffusion inference, one-step diffusion (OSD) methods~\cite{wu2024osediff,wang2024sinsr,gong2025haodiff} significantly reduce inference time. OSD methods can maintain restoration quality and provide a new technical pathway for JPEG compression artifacts removal.

Nevertheless, the inherent complexity of JPEG compression stems from its variable quality factor (QF) and diverse quantization table specifications, leading to diversity in degradation patterns. Previous studies~\cite{guo2025codiff} achieve good results by strategically incorporating QF prior information into the guidance prompts of diffusion models. As mentioned earlier, pre-trained T2I models can achieve optimal performance under text information guidance. On this basis, using textual descriptions as auxiliary information to fine-tune the diffusion process for image restoration proves to be effective~\cite{wu2024osediff,wu2024seesr,yu2024supir}. Existing methods commonly employ additional vision-language models to generate text prompts online during inference. However, such practice poses two critical problems: (1) The rich information contained in low-quality (LQ) images is significantly diluted during conversion to text prompts. (2) Inference efficiency is constrained by the performance of pre-trained vision-language models.

To address the aforementioned issues, we propose SODiff, a novel semantic-oriented one-step diffusion (OSD) model for JPEG compression artifacts removal. At its core is the semantic-aligned image prompt extractor (SAIPE), which employs a Swin Transformer architecture-based image feature extractor to obtain guidance information. In generating embedded ``image prompts", SAIPE achieves alignment with textual description information embedded by the text encoder. Our core motivation is that pre-trained T2I models possess optimal generation prior knowledge. This prior performs best under semantic guidance. SAIPE can utilize the powerful generation prior capabilities of pre-trained models by aligning text and image embeddings.

Furthermore, for LQ images with different compression degrees, the extent of texture detail performs differently from visual information loss. The denoising process at different timesteps in the pre-trained diffusion model corresponds to blocked semantic structures at early steps and refined details at later steps. We propose a quality factor-aware timestep predictor that implicitly learns the QF of LQ images. Such a predictor can select appropriate timesteps for injecting into the diffusion UNet during the denoising process to achieve better restoration effects.

In general, our contributions can be summarized as:
\begin{itemize}
    \item We propose SODiff, a novel OSD model for JPEG compression artifacts removal, which explores the guiding role of rich embedded visual features in the diffusion process at the textual semantic level.
    \item We design a semantic-aligned image prompt extractor (SAIPE) that can extract rich information from LQ images to guide the diffusion process, enabling it to extract ``distilled semantic guidance" while preserving the feature priors of the images themselves.
    \item We design a quality factor (QF)-aware time predictor that attempts to use compression QF as a timestep predictor for diffusion models, selecting the most suitable noise for LQ images with different compression degrees.
    \item Our SODiff can reconstruct missing details under severe compression with high fidelity. SODiff outperforms recent leading methods on multiple existing datasets.
\end{itemize}
\vspace{-4mm}

\section{Related Works}
\subsection{JPEG Artifacts Removal}
Over-compressed JPEG files produce noticeable artifacts, and deep learning-based JPEG artifacts removal methods have achieved remarkable success in recent years. ARCNN~\cite{Dong2015ARCNN} is the first attempt in this field to apply super-resolution networks for artifact mitigation. Some Transformer-based methods~\cite{han2024jdec,Zhang2019RNAN}  also demonstrate their effectiveness in this task. GAN network is introduced to this task by some researchers~\cite{zhang2021bsrgan,galteri2019jpeggan}, aiming to improve the perceptual quality of restored images through generative adversarial networks. Meanwhile, to simultaneously utilize the rich information in both pixel and frequency domains, dual-domain convolutional network methods~\cite{guo2016dualdomain,zhang2018dmcnn} emerge. Prior information of JPEG compression is proven to provide strong guidance for the restoration process. The ranker-guided framework~\cite{wang2021jpegranker} utilizes the ranking of compression quality as auxiliary information. For blind removal scenarios where the QF is unknown, FBCNN~\cite{Jiang2021FBCNN} balances artifact removal and detail preservation by predicting an adjustable quality factor. PromptCIR~\cite{Li2024PromptCIR} explores the application of prompt learning for this task. Based on the powerful image generation priors of pre-trained T2I models, CODiff~\cite{guo2025codiff} achieves excellent results by implicitly learning the QF of LQ images and guiding detail recovery within OSD.

\vspace{-3mm}
\subsection{Diffusion Models}
\vspace{-1mm}
Pre-trained text-to-image (T2I) diffusion models possess powerful natural image generation capabilities through conditional guided denoising processes in the latent space. Some methods incorporate degraded images into the diffusion process~\cite{lin2024diffbir,yu2024supir,wu2024seesr}, achieving excellent image restoration effects by fine-tuning pre-trained networks. DiffBIR~\cite{lin2024diffbir} uses ControlNet to incorporate high-fidelity images as guidance information to control the denoising process, providing a classic paradigm for multi-step diffusion blind denoising. SUPIR~\cite{yu2024supir} employs SDXL~\cite{podell2023sdxl} to generate high-resolution restoration results with impressive amounts of details. Different from the strategy of sampling from pure noise, one-step denoising models ~\cite{wu2024osediff,wang2024sinsr,2024s3diff}use low-quality (LQ) images encoded into latent space as the diffusion starting point, achieving restoration effects through a one-step denoising process. This approach avoids the lengthy inference time and large parameter count of multi-step diffusion. Among them, OSEDiff~\cite{wu2024osediff} aligns the denoising process through variational score distillation, while S3Diff~\cite{2024s3diff} performs diffusion by identifying degradation information to inject LoRA fine-tuning. These image restoration methods provide reference paradigms for severe JPEG compression artifacts removal, but additional task-specific problems need to be considered.
% 预训练的Text to Image（T2I）Diffusion模型，通过在latent space内进行条件引导的去噪过程，具有强大的自然图像生成能力。一些方法将退化后的图像参与进扩散过程~\cite{},通过fine tune预训练的网络，取得了卓越的图像复原效果。DiffBIR~\cite{}通过control-net将高保真图像作为引导信息控制去噪过程，提供了多步扩散盲去噪点经典范式。SUPIR~\cite{}使用SDXL~\cite{}能够生成伴有impressive 细节量的高分辨率复原结果。不同于从纯噪声开始采样的策略，单步去噪模型将编码进入latent space的Low Quality（LQ）图像作为扩散起点，通过一步去噪的过程达到复原的效果。这样的方式避免了多步扩散漫长的推理时间和庞大的参数量。其中，OSEDiff~\cite{}通过变分分数蒸馏进行去噪过程的对齐，而S3Diff通过识别LQ图像退化信息注入Lora微调来进行扩散。这些图像复原方法为严重的JPEG压缩伪影去除提供了参考范式，但是将其介绍道到此任务时需要考虑额外的任务先验。

\begin{figure*}[t]
\begin{center}
\includegraphics[width=0.98\textwidth]{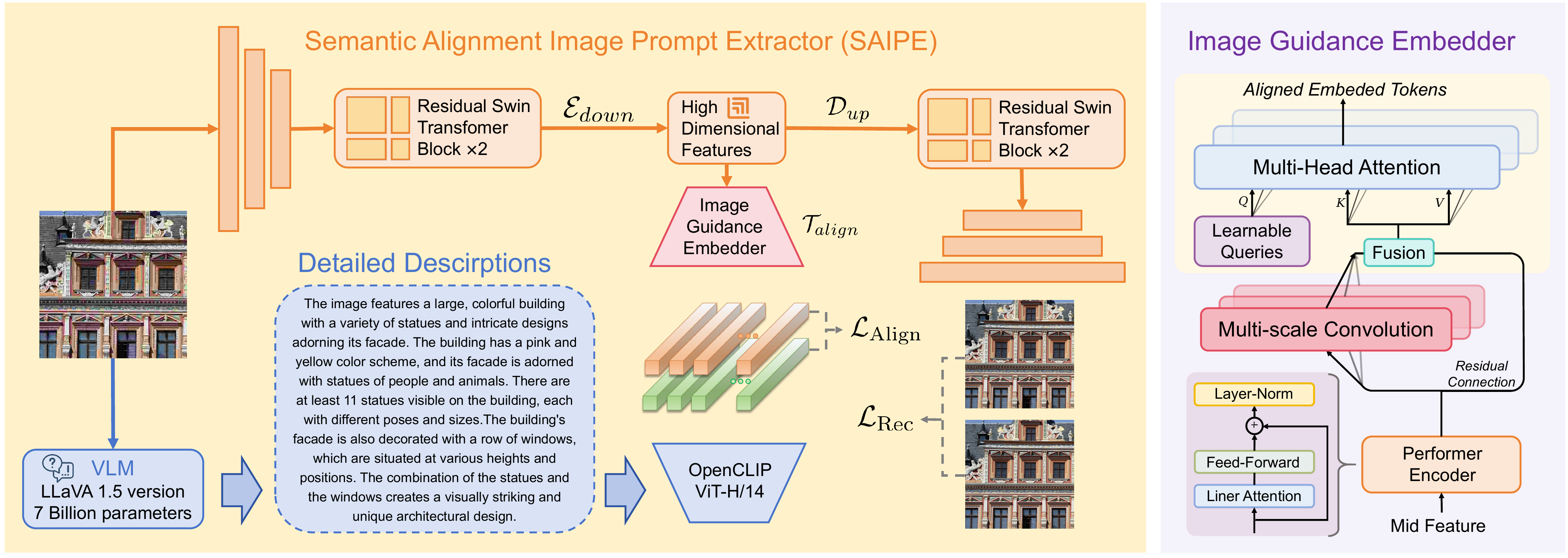}
\end{center}
\vspace{-3mm}
\caption{\textbf{Training Framework of SAIPE}. SAIPE employs a dual-branch training approach with a shared feature extractor $\mathcal{E}_{down}$. The LQ input image $I_L$ is processed to generate intermediate features $F_{mid}$, which feed into two branches. The reconstruction decoder $\mathcal{D}_{up}$ outputs $\hat{I}_{rec}$ and is optimized via reconstruction loss $\mathcal{L}_{rec}$ against the high-quality ground truth $I_H$. The image guidance embedder $\mathcal{T}_{align}$ produces semantic image embeddings $\text{e}_{img}$. To strengthen semantic associations, LLaVA-v1.5-7B generates descriptive text from $I_L$, which is embedded to obtain $\text{e}_{text}$. The alignment loss $\mathcal{L}_{align}$ enforces semantic consistency between $\text{e}_{img}$ and $\text{e}_{text}$ in the embedding space, enabling effective guidance for pre-trained T2I diffusion models.
}
\vspace{-5mm}
\label{fig:saipe}
\end{figure*}

\vspace{-3mm}
\section{Methods}

\subsection{Semantic-Aligned Image Prompt Extractor}
\label{sec:saipe}
Previous research~\cite {wang2024osdface,gong2025osdhuman} demonstrates the effectiveness of extracting image features as guidance information for image restoration. To guide pre-trained models more effectively, the extracted visual features should retain richer reconstruction information than pure text. To this end, we first train a semantic-aligned image prompt extractor (SAIPE). In the one-step diffusion stage, SAIPE can extract key image prompts from compressed low-quality image $I_{L}$ to guide the diffusion process.

\textbf{Model Architecture}. As shown in Fig.~\ref{fig:saipe}, the main structure of SAIPE consists of a shared feature extractor encoder $\mathcal{E}_{down}$, image guidance embedder $\mathcal{T}_{align}$, and reconstruction decoder $\mathcal{D}_{up}$. Among them, the shared feature extractor encoder $\mathcal{E}_{down}$ is designed based on the SwinIR~\cite{liang2021swinir} architecture. The input 512$\times$512 compressed image $I_{L}$ first undergoes 4$\times$ downsampling and expands the channels to 180. It subsequently passes through two layers of residual Swin Transformer blocks (RSTB). It can be detailed as being processed through Swin Transformer layers (STL) with skip connections to preserve shallow-layer information. Through this process, it outputs intermediate feature representations $F_{mid}$ containing rich hierarchical information. $\mathcal{D}_{up}$ correspondingly adopts a symmetric structure with reduced application of STL. After performing Layer Norm, upsampling is conducted to obtain the predicted restored image $\hat{I}_{rec}$. This process can be represented as:
\begin{equation}
F_{mid} = \mathcal{E}_{down}(I_{L}) \quad \text{with} \quad
\hat{I}_{rec} = \mathcal{D}_{up}(F_{mid}).
\end{equation}

Although high-dimensional features $F_{mid}$ contain rich visual information, they cannot directly serve as semantic information to guide the diffusion process. Since the most suitable guidance information for pre-trained T2I models is text, designing an adapter that can focus on the semantic information of image features and establish strong associations with text semantic information is crucial. This approach can be analogized to the process of embedding ``image prompts", for which we design an image guidance embedder. After $F_{mid}$ passes through an input MLP layer, it is processed through a Performer~\cite{choromanski2021performer} encoder, obtaining refined features. Following multi-scale convolutions, these serve as keys and values that enter the multi-head attention pooling layer, where they interact with learnable queries to produce image embeddings  $\text{e}_{img}$ matching the text encoder. This process can be expressed as:
\begin{equation}
   \text{e}_{img} = \mathcal{T}_{align}(F_{mid}).
\end{equation}

\textbf{Training Objective}. During training, we need to preserve the original reconstruction guidance while performing semantic feature alignment. The purpose of this is to remove the degradation information obtained during the compression process of $I_{L}$. Therefore, for the reconstructed image $\hat{I}_{rec}$ after $\mathcal{D}_{up}$, we need to compute the reconstruction loss with the non-compressed image $I_{H}$:
\begin{equation}
\mathcal{L}_{rec} = \mathcal{L}_1(\hat{I}_{rec}, I_{H} ).
\end{equation}

Although the compressed LQ image $I_{L}$ has suboptimal image quality, it is still sufficient to guide highly robust large-scale Vision-Language Models (VLM) in generating detailed textual descriptions. Here, to strengthen the semantic features within the image embeddings, we employ LLaVA-v1.5-7B~\cite{liu2023llava} to generate corresponding descriptive text from $I_{L}$. These detailed descriptions are embedded through the text embedder of the pre-trained T2I model to obtain $\text{e}_{text}$. To align $\text{e}_{img}$ and $\text{e}_{text}$ in the embedding space, we compute the MSE loss between them, allowing the semantically relevant parts within the image embeddings to be as close as possible to the spatial structure of the text embeddings. Therefore, the alignment loss $\mathcal{L}_{align}$ and the total training loss $\mathcal{L}$ can be expressed as:
\begin{gather}
    \mathcal{L}_{align} = \mathcal{L}_\text{MSE}(\text{e}_{img},\text{e}_{text}),  \\
    \mathcal{L} = \mathcal{L}_{rec} + \lambda_{align}\cdot\mathcal{L}_{align}.  
    \label{eq:saipe_loss}
\end{gather}

\begin{figure}[t]
    \centering 
    
    % 第一个子图
    \begin{subfigure}[b]{0.48\columnwidth} 
        \centering
        \includegraphics[width=\textwidth]{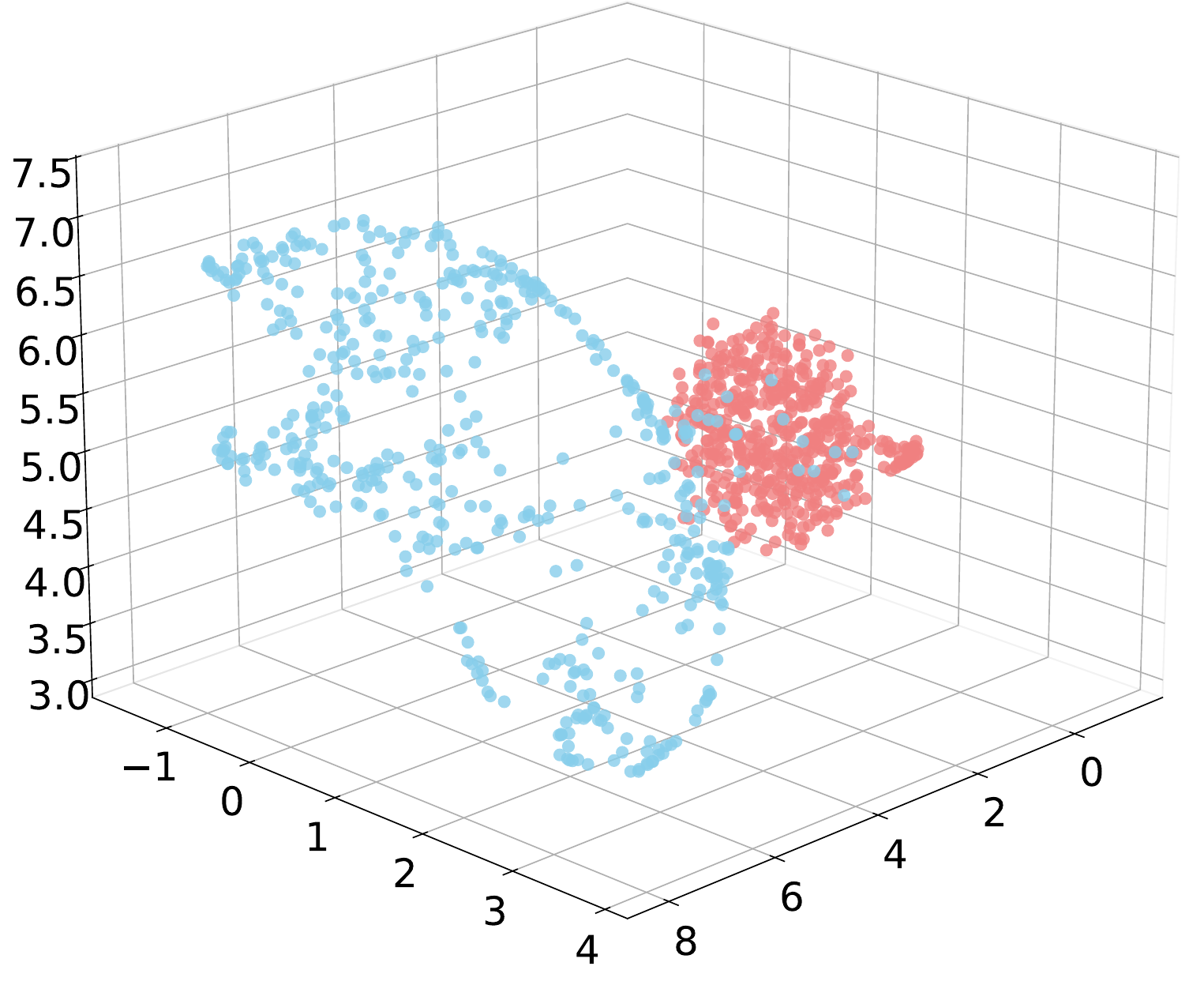}
        \caption{w/o alignment}
        \label{fig:sub_a}
    \end{subfigure}
    \hfill
    % 第二个子图
    \begin{subfigure}[b]{0.48\columnwidth}
        \centering
        \includegraphics[width=\textwidth]{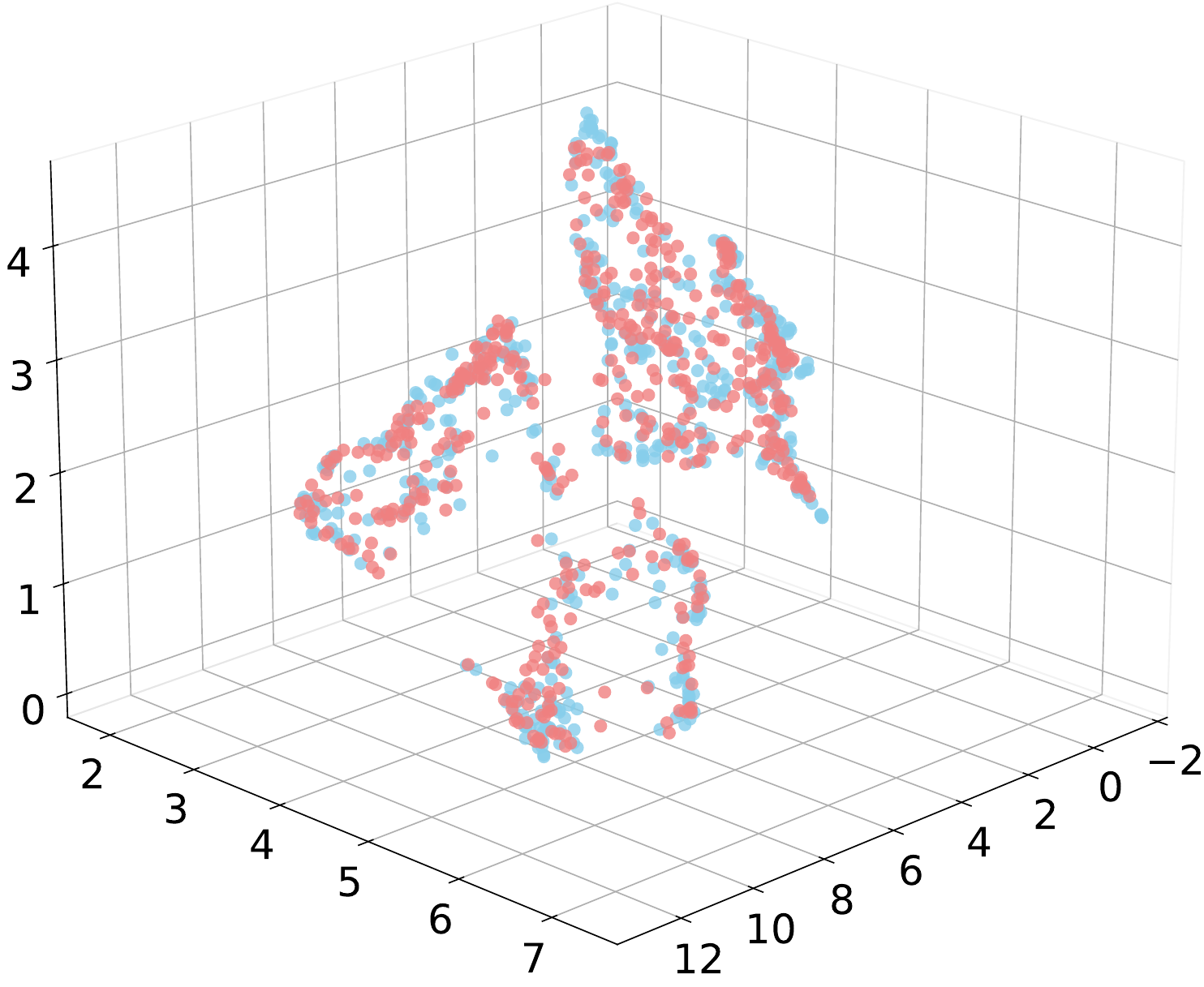}
        \caption{w/ alignment}
        \label{fig:sub_b}
    \end{subfigure}
    \vspace{-3mm}
    \caption{\textbf{Visualization of the effectiveness of semantic-aligned training objective strategy}. \textbf{Red} dots represent text embeddings, while \textbf{blue} dots represent aligned embeddings under different $\mathcal{T}_{align}$ training strategies. The results under the alignment training strategy are significantly closer to the distribution of text embeddings. Zoom in for better view.}
    \label{fig:umap}
    \vspace{-6mm}
\end{figure}

To intuitively illustrate that SAIPE enhances the parts with stronger semantic associations in image embeddings, we performed Uniform Manifold Approximation and Projection (UMAP) ~\cite{mcinnes2018umap} dimensionality reduction on the embeddings. For comparison, we first train the reconstruction branch according to the same strategy, then train $\mathcal{T}_{align}$ together with the Diffusion model. Fig.~\ref{fig:umap} shows their comparison results. It can be observed that in the high-dimensional space distribution, $\mathcal{T}_{align}$ constrained by alignment loss is closer to the magnitude and spatial distribution of text embeddings. For a detailed comparison of model performance, please refer to the ablation experiments in Section~\ref{sec:ablation}.

\begin{figure*}[t]
\begin{center}
\includegraphics[width=\textwidth]{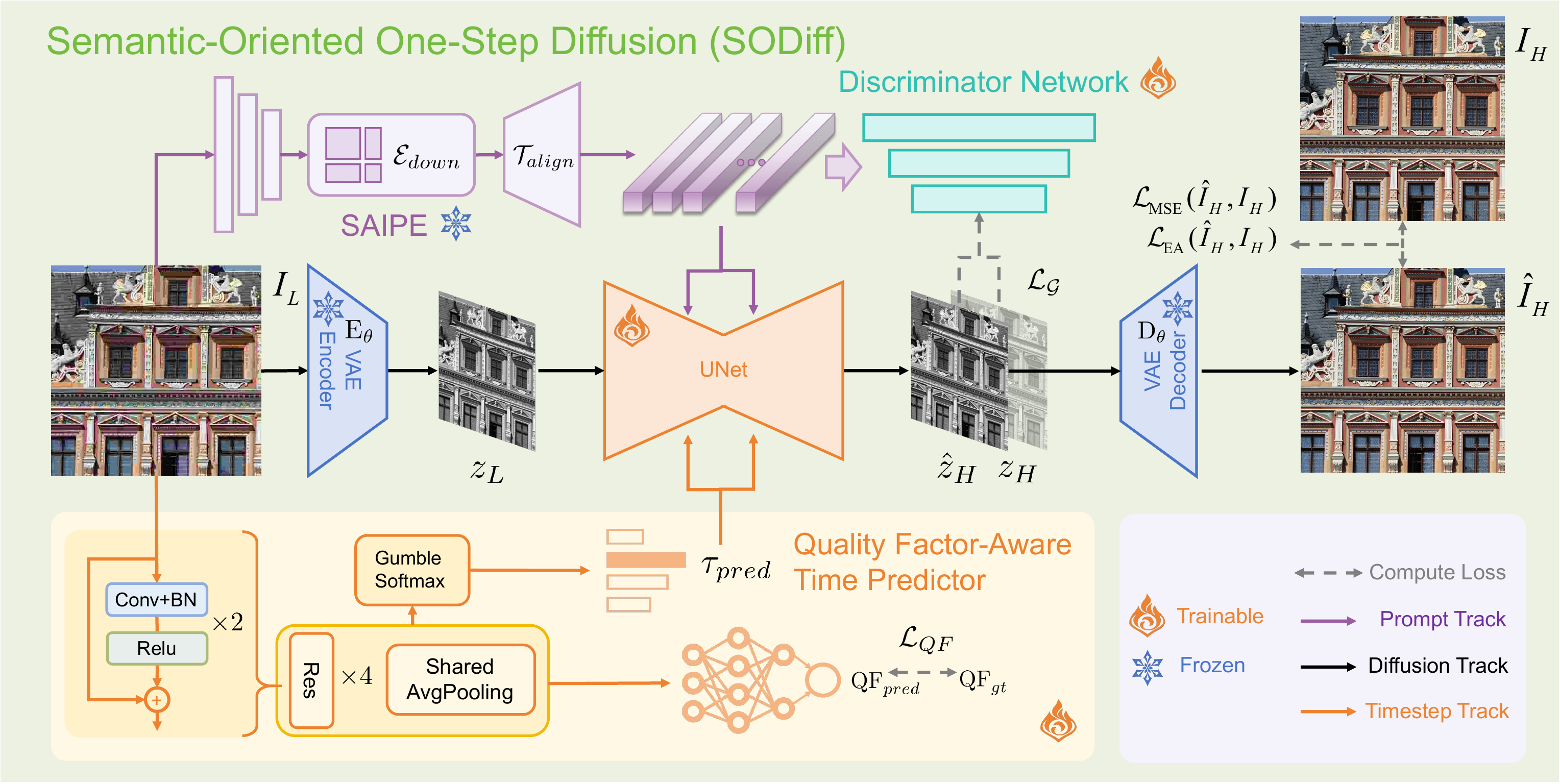}
\end{center}
\vspace{-5mm}
\caption{Training Framework of SODiff. \textbf{First}, the Prompt Track uses the first-stage trained SAIPE to extract semantic-oriented embedded image guidance $e_{img}$ from $I_L$. \textbf{Second}, the Timestep Track performs OSD timestep and QF prediction, obtaining $\tau_{pred}$ and computing $\mathcal{L}_{qf}$. \textbf{Next}, $I_L$ is encoded into latent space through the frozen VAE encoder to obtain $z_L$, with $e_{img}$ and $\tau_{pred}$ providing conditional control for the noise prediction UNet, which performs one-step noise prediction to obtain $\hat{z}_H$. The SDXL Discriminator receives the transformed $e_{img}$, $\hat{z}_H$, and the latent space representation $z_H$ of $I_H$ to compute $\mathcal{L}_\mathcal{G}$. \textbf{Last}, $\hat{z}_H$ passes through the frozen VAE Decoder to obtain the reconstructed image $\hat{I}_H$, which computes $\mathcal{L}_{recon}$ with $I_H$.
}

\vspace{-6mm}
\label{fig:sodiff}
\end{figure*}

\vspace{-2mm}
\subsection{One-Step Diffusion (OSD) Model}
% \vspace{-2mm}

\textbf{Model Architecture.} Recently, diffusion-based image restoration research~\cite{wu2024osediff,wang2024osdface,gong2025osdhuman}, many approaches leverage pretrained text-to-image (T2I) models to incorporate strong generative priors. In our method, we adopt Stable Diffusion (SD)~\cite{sd21} as the base model and retain the original architecture of both the variational autoencoder (VAE)~\cite{kingma2014vae} and the UNet~\cite{ronneberger2015unet} components. After obtaining the latent vector $z_L = E_\theta(I_L)$ through the VAE encoder $E_\theta$, our model follows a denoising process similar to that in SD to predict the high-quality latent vector $\hat{z}_H$:
\vspace{-1mm}
\begin{equation} 
\hat{z}_H =\frac{z_L - \sqrt{1 - \bar{\alpha}_{\tau}} \varepsilon_{\theta} (z_L; \text{e}_{img},\tau_{pred})}{\sqrt{\bar{\alpha}_{\tau}}},
\label{eq:generate_zh} 
\end{equation}
where $\varepsilon_\theta$ denotes the denoising UNet guided by both the prompt embedding $\text{e}_{img}$ and the timestep $\tau_{pred}$. Specifically, $\text{e}_{img}$ is the output of the image guidance embedder of the semantic-aligned image prompt extractor (SAIPE), while $\tau_{pred}$ is predicted by the quality factor-aware time predictor. Finally, the restored high-quality image $\hat{I}_H$ is obtained by decoding $\hat{z}_H$ using the VAE decoder $D_\theta$: $\hat{I}_H = D_\theta(\hat{z}_H)$.

\textbf{Training Objective}. During training, we employ three types of loss functions: reconstruction loss, adversarial loss, and quality factor (QF)-aware loss, which is mentioned in Sec.~\ref{sec:timestep}. The reconstruction loss consists of an MSE term and an edge-aware DISTS perceptual loss which enhances the sensitivity of the DISTS~\cite{ding2020dists} metric to image edges using the Sobel operator $\mathcal{S}(\cdot)$. Therefore, the reconstruction loss $\mathcal{L}_{\text{recon}}$ can be formulated as:
\begin{equation}
\begin{aligned}
\mathcal{L}_{\text{recon}} =& \mathcal{L}_{\text{MSE}}(\hat{I}_H, I_H) + \mathcal{L}_\text{EA} \quad \text{with} \\
\mathcal{L}_\text{EA} =\mathcal{L}&_{\text{DISTS}}(\mathcal{S}(\hat{I}_H), \mathcal{S}(I_H)) + \mathcal{L}_{\text{DISTS}}(\hat{I}_H, I_H).     
\end{aligned}
\end{equation}
Meanwhile, previous work~\cite{li2025d3sr} reveals a significant discrepancy between predicted and ground truth distributions in the latent space, which reconstruction losses in the RGB domain alone cannot effectively mitigate. Following this line of research, we adopt a pre-trained diffusion UNet as the discriminator $\mathcal{D}$. The loss functions for the generator $\mathcal{G}$ and discriminator $\mathcal{D}$ are defined as follows:
\begin{equation}
\mathcal{L}_\mathcal{G}(\hat{z}_H) = -\mathbb{E}_t \left[\log \mathcal{D} \left( F(\hat{z}_H, t) \right)\right], 
\end{equation}
\begin{equation}
\begin{aligned}
\mathcal{L}_\mathcal{D}(\hat{z}_H,z_H) = &-\mathbb{E}_t \left[\log \left( 1 - \mathcal{D} \left( F(\hat{z}_H, t) \right) \right)\right] \\
&- \mathbb{E}_t \left[\log \mathcal{D} \left( F(z_H, t) \right)\right], 
\end{aligned}
\end{equation}
where $F(\cdot)$ represents the process of adding diffusion noise, which depends on a random timestep $t \in [0, T]$.

\vspace{-2mm}
\subsection{Quality Factor-Aware Time Predictor}
% \vspace{-4mm}
\label{sec:timestep}
Although compressing the multi-step diffusion process into a single step significantly reduces inference time and computational overhead, it is important to note that the pre-trained T2I Model is originally trained to generate realistic natural images through progressive refinement across multiple diffusion steps~\cite{Ho2020DDPM,song2021ddim}. For a powerful pre-trained SD-UNet, the predicted noise levels at different timesteps in the latent space are not uniform~\cite{Luo2023timestep}. Previous study~\cite{Wang2024HeroSR} demonstrates that introducing a suitable timestep predictor for the OSD model can yield better performance compared to inference with a fixed timestep.

In the context of JPEG compression, a specific quality factor (QF) is typically specified before compression, where a lower QF generally leads to more severe detail loss. To leverage the prior knowledge embedded in the pre-trained model’s noise schedule, we propose a quality factor-aware time predictor. It infers a suitable timestep for the one-step denoising process by analyzing visual features extracted from the compressed LQ image $I_{L}$ via four residual blocks. To further inject JPEG compression priors into the decision process, an additional linear branch following a shared average pooling layer is employed to predict the QF of $I_{L}$. The predicted quality factor $\text{QF}_{pred}$ is supervised using an L1 loss against the ground truth $\text{QF}_{gt}$, formulated as:
\begin{equation}
    \mathcal{L}_{qf} = || \text{QF}_{pred} - \text{QF}_{gt} ||_1.
\end{equation}

For the dedicated timestep prediction branch, ensuring the differentiable property of input timestep is crucial for end-to-end training of the overall Diffusion framework. Since discrete timestep selection would break the gradient flow, we employ a continuous relaxation approach. The predicted logits $l_i, i \in \{0, 1, 2, \ldots, T_{\max}-1\}$ corresponding to different timesteps need to be processed through the Gumbel-Softmax technique~\cite{Jang2017gumble}, which provides a differentiable approximation to categorical sampling. This approach introduces Gumbel noise $g_i \sim \text{Gumbel}(0, 1)$ to the logits before applying softmax normalization, enabling smooth gradient computation while maintaining the stochastic nature of timestep selection. Therefore, the predicted timestep $\tau_{\text{pred}}$ can be expressed as a learnable weighted combination of all possible timesteps:
\begin{equation}
    \tau_{pred} = \sum_{i=0}^{T_{max}-1} i \cdot \frac{\exp(l_i + g_i)}{\sum_{j=0}^{T_{max}-1} \exp((l_j + g_j) }. 
\end{equation}
    
Thus, the overall training loss $\mathcal{L}_{\text{total}}$ is formulated as:
\begin{equation}
\mathcal{L}_{\text{total}} = \mathcal{L}_{\text{recon}}(\hat{I}_H , I_H) + \alpha \cdot\mathcal{L}_\mathcal{G}(\hat{z}_H) + \beta \cdot \mathcal{L}_{qf}.
\label{eq:total_loss}
\end{equation}

\section{Experiments}
\subsection{Experimental Settings}
% \vspace{-1mm}
\textbf{Training and Testing Datasets}.
Our model is trained on the DF2K~\cite{agustsson2017div2k,8014883flick}, LSDIR~\cite{Li2023LSDIR} datasets. For preprocessing, we randomly crop DF2K and LSDIR to a size of 512$\times$512. We randomly select JPEG compression with quality factors in the range [5,95] as the degradation pipeline. For the test set, we follow previous work~\cite{guo2025codiff} and select LIVE-1~\cite{sheikh2005live}, Urban100~\cite{Huang-CVPR-2015urban100}, and DIV2K-val~\cite{agustsson2017div2k}, conducting evaluations at QF values of 5, 10, and 20, respectively.

\textbf{Evaluation Metrics}.
To comprehensively evaluate the model's performance, we selected both full-reference and no-reference image quality assessment (IQA) to assess the test results. The full-reference IQA includes LPIPS~\cite{zhang2018lpips} and DISTS~\cite{ding2020dists}, which can respectively reflect visual perceptual similarity and structural distortion characteristics between $\hat{I}_H$ and $I_H$. The no-reference IQA are MUSIQ~\cite{ke2021musiq}, MANIQA (abbreviated as M-IQA)~\cite{yang2022maniqa}, and CLIPIQA (abbreviated as C-IQA)~\cite{wang2022clipiqa}, which are used to provide quality evaluation that aligns with natural human perceptual judgment.

\textbf{Implementation Details}.
The overall training process consists of two stages: the training of SAIPE and SODiff. In the first stage, all text descriptions are pre-generated before the training stage, and they are generated from image content within the cropped regions. With $\lambda_{align}$ of alignment loss in Eq.~\eqref{eq:saipe_loss} set to 0.5, we use the Adam~\cite{kingma2017adam} optimizer with a learning rate of 2$\times$10$^{-4}$. In the second stage, we select SD2.1-base~\cite{sd21} as the base model and freeze SAIPE and the VAE encoder and decoder. The LoRA rank is set to 16 to finetune SD. We use AdamW as the optimizer with learning rate of 1$\times$10$^{-5}$. Following D$^3$SR~\cite{li2025d3sr}, we use SDXL as our discriminator network and set $\alpha$ in Eq.~\eqref{eq:total_loss} to 1$\times$10$^{-2}$. $\beta$ in Eq.~\eqref{eq:total_loss} is set to 1$\times$10$^{-3}$. With batch sizes of 16 and 2 for the two stages, training is performed on 4 and 2 NVIDIA RTX A6000 GPUs for 50k and 150k iterations, respectively.

\textbf{Compared State-of-the-Art (SOTA) Methods}
We compare SODiff with two categories of models. First, we compare with non-diffusion models, including FBCNN~\cite{Jiang2021FBCNN}, JDEC~\cite{han2024jdec}, and PromptCIR~\cite{Li2024PromptCIR}. Second, we compare with diffusion-based models, including DiffBIR~\cite{lin2024diffbir} and SUPIR~\cite{yu2024supir}. Note that DiffBIR is retrained for JPEG compression artifacts removal.
\begin{table*}[!ht]
\centering
\small 
\setlength{\tabcolsep}{3pt}
\begin{tabular}{l|ccc|ccc|ccc|ccc|ccc}
\toprule
\multicolumn{16}{c}{\textbf{LIVE-1}} \\
\midrule
\multirow{2}{*}{Method} & \multicolumn{3}{c|}{LPIPS $\downarrow$} & \multicolumn{3}{c|}{DISTS $\downarrow$} & \multicolumn{3}{c|}{MUSIQ $\uparrow$} & \multicolumn{3}{c|}{MANIQA $\uparrow$} & \multicolumn{3}{c}{CLIPIQA $\uparrow$} \\
 & QF5 & QF10 & QF20 & QF5 & QF10 & QF20 & QF5 & QF10 & QF20 & QF5 & QF10 & QF20 & QF5 & QF10 & QF20 \\
\midrule
JPEG & 0.4384 & 0.3013 & 0.1799 & 0.3242 & 0.2387 & 0.1653 & 40.33 & 53.88 & 64.12 & 0.2294 & 0.3509 & 0.4411 & 0.1716 & 0.2737 & 0.5542 \\
FBCNN & 0.3736 & 0.2503 & 0.1583 & 0.2353 & 0.1785 & 0.1319 & \underline{63.56} & 71.00 & 73.96 & \underline{0.3425} & 0.4207 & 0.4551 & 0.2763 & 0.4767 & 0.5535 \\
JDEC & 0.4113 & 0.2450 & 0.1555 & 0.2364 & 0.1740 & 0.1282 & 55.66 & 70.80 & 73.81 & 0.2002 & 0.4065 & 0.4433 & 0.1539 & 0.4811 & 0.5512 \\
PromptCIR & 0.3797 & 0.2290 & \underline{0.1450} & 0.2334 & 0.1658 & 0.1223 & 60.34 & \underline{72.39} & \textbf{74.12} & 0.2790 & 0.4500 & 0.4713 & 0.2655 & 0.5176 & 0.5847 \\
DiffBIR* & \underline{0.3509} & \underline{0.2160} & 0.1500 & \underline{0.2035} & \underline{0.1319} & \underline{0.0988} & 58.09 & 67.38 & 71.08 & 0.2812 & 0.3789 & 0.4371 & \underline{0.3776} & 0.5789 & 0.6814 \\
SUPIR & 0.4856 & 0.2770 & 0.1683 & 0.2720 & 0.1558 & 0.1121 & 52.69 & 68.77 & 73.02 & 0.3229 & \underline{0.5183} & \textbf{0.6237} & 0.3149 & \underline{0.6115} & \underline{0.7364} \\
\midrule
SODiff (ours) & \textbf{0.2229} & \textbf{0.1605} & \textbf{0.1237} & \textbf{0.1173} & \textbf{0.0938} & \textbf{0.0763} & \textbf{72.88} & \textbf{73.84} & \underline{74.11} & \textbf{0.4957} & \textbf{0.5192} & \underline{0.5272} & \textbf{0.7087} & \textbf{0.7323} & \textbf{0.7587} \\
\midrule
\multicolumn{16}{c}{\textbf{Urban100}} \\
\midrule
\multirow{2}{*}{Method} & \multicolumn{3}{c|}{LPIPS $\downarrow$} & \multicolumn{3}{c|}{DISTS $\downarrow$} & \multicolumn{3}{c|}{MUSIQ $\uparrow$} & \multicolumn{3}{c|}{MANIQA $\uparrow$} & \multicolumn{3}{c}{CLIPIQA $\uparrow$} \\
 & QF5 & QF10 & QF20 & QF5 & QF10 & QF20 & QF5 & QF10 & QF20 & QF5 & QF10 & QF20 & QF5 & QF10 & QF20 \\
\midrule
JPEG & 0.3481 & 0.2254 & 0.1244 & 0.2834 & 0.2145 & 0.1521 & 50.46 & 60.87 & 67.60 & 0.3656 & 0.4401 & 0.4967 & 0.2806 & 0.3517 & 0.5343 \\
FBCNN & 0.2341 & 0.1462 & 0.0896 & 0.2162 & 0.1648 & 0.1249 & 69.03 & \underline{72.55} & \underline{73.39} & 0.4263 & 0.5033 & 0.5288 & 0.3800 & 0.5014 & 0.5437 \\
JDEC & 0.2794 & 0.1382 & \underline{0.0846} & 0.2309 & 0.1570 & 0.1175 & 62.97 & 72.52 & 73.30 & 0.3386 & 0.5001 & 0.5230 & 0.2518 & 0.4959 & 0.5369 \\
PromptCIR & 0.2389 & \underline{0.1183} & \textbf{0.0739} & 0.2037 & 0.1431 & 0.1083 & 66.08 & \textbf{73.01} & \textbf{73.47} & 0.3946 & 0.5380 & 0.5489 & 0.3619 & 0.5337 & 0.5662 \\
DiffBIR* & \underline{0.2018} & 0.1344 & 0.1005 & \underline{0.1657} & \underline{0.1207} & \underline{0.0939} & 69.63 & 71.77 & 72.51 & 0.4285 & 0.4813 & 0.5105 & 0.5470 & 0.5966 & 0.6306 \\
SUPIR & 0.3279 & 0.2489 & 0.2125 & 0.2018 & 0.1659 & 0.1518 & \underline{69.94} & 72.37 & 73.01 & \underline{0.5546} & \textbf{0.5995} & \textbf{0.6105} & \underline{0.5536} & \underline{0.6178} & \underline{0.6397} \\
\midrule
SODiff (ours) & \textbf{0.1579} & \textbf{0.1098} & \underline{0.0846} & \textbf{0.1196} & \textbf{0.0914} & \textbf{0.0734} & \textbf{71.33} & 72.23 & 72.63 & \textbf{0.5598} & \underline{0.5451} & \underline{0.5561} & \textbf{0.6392} & \textbf{0.6558} & \textbf{0.6733} \\
\midrule
\multicolumn{16}{c}{\textbf{DIV2K-val}} \\
\midrule
\multirow{2}{*}{Method} & \multicolumn{3}{c|}{LPIPS $\downarrow$} & \multicolumn{3}{c|}{DISTS $\downarrow$} & \multicolumn{3}{c|}{MUSIQ $\uparrow$} & \multicolumn{3}{c|}{MANIQA $\uparrow$} & \multicolumn{3}{c}{CLIPIQA $\uparrow$} \\
 & QF5 & QF10 & QF20 & QF5 & QF10 & QF20 & QF5 & QF10 & QF20 & QF5 & QF10 & QF20 & QF5 & QF10 & QF20 \\
\midrule
JPEG & 0.3459 & 0.3234 & 0.2072 & 0.2570 & 0.2255 & 0.1465 & 25.95 & 47.53 & 57.45 & 0.2570 & 0.3120 & 0.3557 & 0.2595 & 0.3303 & 0.5072 \\
FBCNN & 0.3445 & 0.2448 & 0.1733 & 0.2078 & 0.1581 & 0.1168 & 56.52 & 61.79 & 65.20 & 0.3025 & 0.3593 & 0.3775 & 0.3004 & 0.4561 & 0.5221 \\
JDEC & 0.3811 & 0.2313 & 0.1565 & 0.2234 & 0.1574 & 0.1152 & 53.88 & \textbf{67.48} & \textbf{69.90} & 0.2118 & 0.3689 & 0.3927 & 0.1841 & 0.4675 & 0.5319 \\
PromptCIR & 0.3549 & 0.2240 & 0.1581 & 0.2067 & 0.1459 & 0.1061 & 52.21 & 62.63 & 65.62 & 0.2705 & \underline{0.3758} & 0.3871 & 0.3041 & 0.4956 & 0.5483 \\
DiffBIR* & \underline{0.2788} & \underline{0.1953} & \underline{0.1542} & \underline{0.1533} & \underline{0.1072} & \underline{0.0856} & \underline{60.21} & 65.22 & \underline{67.06} & 0.3220 & 0.3754 & \textbf{0.4033} & \underline{0.4975} & \underline{0.5912} & \underline{0.6355} \\
SUPIR & 0.4372 & 0.3121 & 0.2295 & 0.2148 & 0.1410 & 0.1161 & 54.07 & 61.93 & 64.87 & \underline{0.3438} & 0.3570 & 0.3723 & 0.4219 & 0.5186 & 0.5535 \\
\midrule
SODiff (ours) & \textbf{0.2425} & \textbf{0.1732} & \textbf{0.1295} & \textbf{0.1126} & \textbf{0.0816} & \textbf{0.0622} & \textbf{64.42} & \underline{65.90} & 66.49 & \textbf{0.3733} & \textbf{0.3924} & \underline{0.3984} & \textbf{0.5851} & \textbf{0.6193} & \textbf{0.6398} \\
\bottomrule
\end{tabular}
\vspace{-2mm}
\caption{Performance comparisons across datasets and different QF. The best and second-best results are shown in \textbf{bold} and \underline{underlined}, respectively. Model with asterisk (*) denotes that it is retrained for JPEG compression artifacts removal.}
\label{tab:results}
\vspace{-3mm}
\end{table*}

% \vspace{-4mm}
\subsection{Main Results}
% \vspace{-1mm}

\textbf{Quantitative Comparisons.} Table~\ref{tab:results} presents the quantitative comparison results across three benchmark datasets (LIVE-1, Urban100, and DIV2K-val) under different quality factors (QF). SODiff consistently outperforms all baseline methods, including traditional non-diffusion approaches (FBCNN, JDEC, PromptCIR) and existing diffusion-based methods (DiffBIR, SUPIR). Our method achieves the best results in perceptual quality metrics and competitive performance in no-reference quality measures across all evaluation scenarios. The performance advantage becomes more pronounced at lower quality factors, indicating SODiff's robustness in handling severe compression artifacts and validating the effectiveness of our semantic-oriented approach.

\begin{table}[t]
    \centering
    \small
    \setlength{\tabcolsep}{1pt}
    \begin{tabular}{l|ccc|ccc}
        \toprule
        & \multicolumn{3}{c|}{Urban100} & \multicolumn{3}{c}{DIV2K-Val} \\
        Method & DISTS$\downarrow$ & MUSIQ$\uparrow$ & M-IQA$\uparrow$ & DISTS$\downarrow$ & MUSIQ$\uparrow$ & M-IQA$\uparrow$ \\
        \midrule
        w/o $\mathcal{L}_{align}$ & 0.1261 & 64.41 & 0.4609 & 0.1071 & 62.83 & 0.3250 \\
        DAPE & 0.0877 & {71.53} & {0.4953} & \textbf{0.0697} & {64.39} & {0.3652} \\
        \midrule
        SAIPE & \textbf{0.0862} & \textbf{72.51} & \textbf{0.5531} & 0.0731 & \textbf{66.72} & \textbf{0.4013} \\
        \bottomrule
    \end{tabular}
    \vspace{-2mm}
    \caption{Ablation study of prompt methods on Urban100 and DIV2K-val datasets. The best results are shown in \textbf{bold}.} 
    \label{table:embed}
    \vspace{-6mm}
\end{table}

\begin{figure*}[t]
\centering

% 第一组 - 第一行
\begin{subfigure}[t]{0.24\textwidth}
\centering
\includegraphics[width=\textwidth]{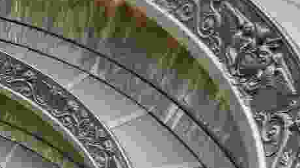}  
\caption{LQ (Urban100: img\_051)}
\label{fig:1a}
\end{subfigure}
\hfill
\begin{subfigure}[t]{0.24\textwidth}
\centering
\includegraphics[width=\textwidth]{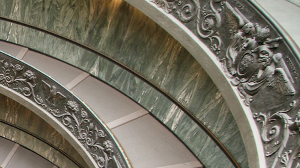}  
\caption{HQ}
\label{fig:1b}
\end{subfigure}
\hfill
\begin{subfigure}[t]{0.24\textwidth}
\centering
\includegraphics[width=\textwidth]{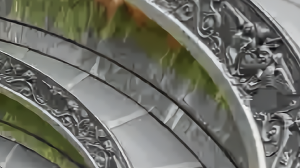}  
\caption{FBCNN}
\label{fig:1c}
\end{subfigure}
\hfill
\begin{subfigure}[t]{0.24\textwidth}
\centering
\includegraphics[width=\textwidth]{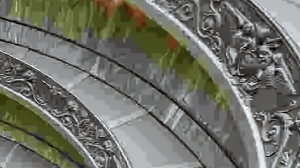}  
\caption{JDEC}
\label{fig:1d}
\end{subfigure}

% 第一组 - 第二行
\begin{subfigure}[t]{0.24\textwidth}
\centering
\includegraphics[width=\textwidth]{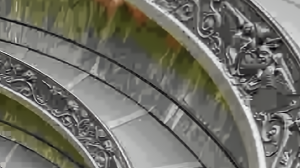}  
\caption{PromptCIR}
\label{fig:1e}
\end{subfigure}
\hfill
\begin{subfigure}[t]{0.24\textwidth}
\centering
\includegraphics[width=\textwidth]{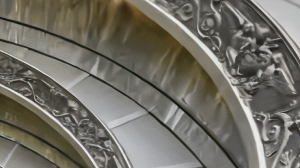}  
\caption{DiffBIR*}
\label{fig:1f}
\end{subfigure}
\hfill
\begin{subfigure}[t]{0.24\textwidth}
\centering
\includegraphics[width=\textwidth]{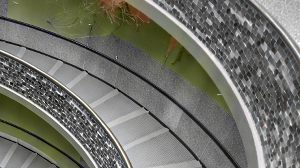}  
\caption{SUPIR}
\label{fig:1g}
\end{subfigure}
\hfill
\begin{subfigure}[t]{0.24\textwidth}
\centering
\includegraphics[width=\textwidth]{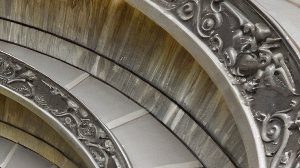}  
\caption{SODiff (ours)}
\label{fig:1h}
\end{subfigure}

% 第二组 - 第一行
\begin{subfigure}[t]{0.24\textwidth}
\centering
\includegraphics[width=\textwidth]{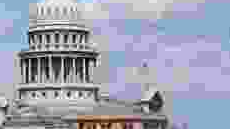}  
\caption{LQ (LIVE-1: church.)}
\label{fig:2a}
\end{subfigure}
\hfill
\begin{subfigure}[t]{0.24\textwidth}
\centering
\includegraphics[width=\textwidth]{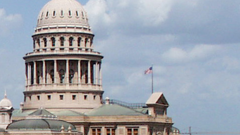}  
\caption{HQ}
\label{fig:2b}
\end{subfigure}
\hfill
\begin{subfigure}[t]{0.24\textwidth}
\centering
\includegraphics[width=\textwidth]{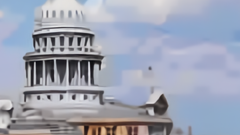}  
\caption{FBCNN}
\label{fig:2c}
\end{subfigure}
\hfill
\begin{subfigure}[t]{0.24\textwidth}
\centering
\includegraphics[width=\textwidth]{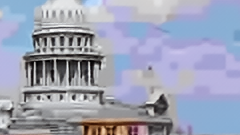}  
\caption{JDEC}
\label{fig:2d}
\end{subfigure}

% 第二组 - 第二行
\begin{subfigure}[t]{0.24\textwidth}
\centering
\includegraphics[width=\textwidth]{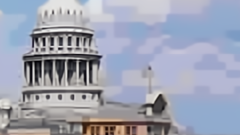}  
\caption{PrompCIR}
\label{fig:2e}
\end{subfigure}
\hfill
\begin{subfigure}[t]{0.24\textwidth}
\centering
\includegraphics[width=\textwidth]{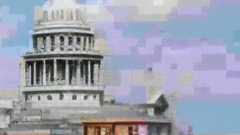}  
\caption{DiffBIR*}
\label{fig:2f}
\end{subfigure}
\hfill
\begin{subfigure}[t]{0.24\textwidth}
\centering
\includegraphics[width=\textwidth]{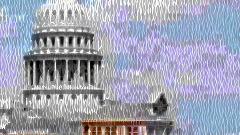}  
\caption{SUPIR}
\label{fig:2g}
\end{subfigure}
\hfill
\begin{subfigure}[t]{0.24\textwidth}
\centering
\includegraphics[width=\textwidth]{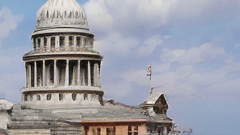}  
\caption{SODiff (ours)}
\label{fig:2h}
\end{subfigure}
\vspace{-2mm}
\caption{Visual comparison (QF=5) on Urban100 and LIVE-1 datasets. Please zoom in for a better view.}
\label{fig:visual}
\vspace{-6mm}
\end{figure*}

% \vspace{-2mm}
\textbf{Qualitative Comparisons.} As illustrated in Fig.~\ref{fig:visual}, when the compressed LQ image severely loses high-frequency details compared to the HQ image, SODiff achieves the best visual results. Non-diffusion architecture methods (FBCNN, JDEC, PromptCIR) perform poorly when facing severe compression artifacts, particularly blurred detail textures and blocky color stratification. These issues are not alleviated in DiffBIR and SUPIR, and these obvious compression artifacts can misleadingly guide the diffusion process, resulting in distorted textures (as shown in Fig.~\ref{fig:1g}). Although DiffBIR is retrained for this task, it still struggles with overly severe color block stratification, and the detail textures in its restoration results are inferior to those of SODiff. SODiff can understand the semantic information in compressed LQ images while focusing on appropriate detail textures, enabling it to restore color blocks with lost structural information into objects that conform to their semantic representations (such as clouds in Fig.~\ref{fig:2h}). More visual comparisons can be found in the supplementary materials.

\vspace{-3mm}

\subsection{Ablation Studies}

\vspace{-1mm}
\label{sec:ablation}
\textbf{The Effectiveness of SAIPE}.
To verify SAIPE's effectiveness, we compare different prompt methods. We employ the degradation-aware prompt extractor (DAPE)~\cite{wu2024seesr} for text extraction comparison rather than LLaVA, which generates more detailed descriptions but has impractical inference time for one-step diffusion. We also test an alternative strategy where only the reconstruction branch was trained first, followed by joint training of $\mathcal{T}_{align}$ with the diffusion model. Results in Tab.~\ref{table:embed} on Urban100 and DIV2K-val datasets demonstrate that SAIPE achieves better restoration guidance than pure text prompts and outperforms image guidance under pure reconstruction guidance.

\begin{table}[t]
    \centering
    % \small
    \setlength{\tabcolsep}{2pt}
    \vspace{2mm}
    \begin{tabular}{l|ccccc} 
        \toprule
        Method & LPIPS$\downarrow$ & DISTS$\downarrow$ & MUSIQ$\uparrow$ & M-IQA$\uparrow$& C-IQA$\uparrow$ \\
        \midrule
        w/o \textbf{TP} & 0.3779 & 0.2577 & 56.1229 & {0.3236} & 0.4105 \\ 
        w/o $\mathcal{L}_{qf}$ & {0.3270} & {0.1611} & {58.7188} & 0.3232 & {0.4616} \\
        \midrule
        Full model & \textbf{0.3124} & \textbf{0.1469} & \textbf{60.3954} & \textbf{0.3292} & \textbf{0.5181} \\
        \bottomrule
    \end{tabular}
    \vspace{-1.5mm}
    \caption{Ablation studies on different components. The best results are shown in \textbf{bold}. \textbf{TP} represents timestep predictor.}
    \label{table:time}
    \vspace{-4mm}
\end{table}

\begin{table}[t]
    \centering
    % \small
    \setlength{\tabcolsep}{2pt}
    \begin{tabular}{cc|ccccc} 
        \toprule
        $\mathcal{L}_{\text{EA}}$ & $\mathcal{L}_\mathcal{G}$ & LPIPS$\downarrow$ & DISTS$\downarrow$ & MUSIQ$\uparrow$ & M-IQA$\uparrow$& C-IQA$\uparrow$ \\
        \midrule
        & \checkmark & {0.3369} & 0.2475 & 56.1691 & 0.3117 & 0.2745 \\
         \checkmark &  & 0.3602 & {0.1776} & {58.1101} & {0.3250} & {0.5022} \\
         \midrule
        \checkmark & \checkmark & \textbf{0.2102} & \textbf{0.1558} & \textbf{69.5945} & \textbf{0.4939} & \textbf{0.5901} \\ 
        \bottomrule
    \end{tabular}
    \vspace{-1.5mm}
    \caption{Ablation studies on different loss functions. The best results are shown in \textbf{bold}.}
    \label{table:loss}
    \vspace{-8mm}
\end{table}

\textbf{Timestep Predictor}.
% 同时，我们验证了是否采用quality factor （QF）aware timestep predictor进行时间步预测对SODiff带来的影响，并且测试了如果没有在训练的时候隐式地学习QF的结果。Table.ref{time}说明在QF aware timestep predictor的帮助下能够辅助模型取得最佳的效果。
We verify the impact of adopting quality factor (QF) aware timestep predictor for timestep prediction on SODiff, and test the results without implicitly learning QF during training. Table~\ref{table:time} demonstrates that with the assistance of QF-aware timestep predictor, the model achieves optimal performance. Moreover, it can be observed that implicit QF learning helps improve all metrics.

\textbf{SODiff Training Loss Functions}.
The adoption of training loss functions is critical for training effectiveness, particularly the edge-aware DISTS loss $\mathcal{L}_\text{EA}$ and adversarial loss $\mathcal{L}_\mathcal{G}$. To evaluate their individual contributions, we conducte experiments by excluding each component. As shown in Tab.~\ref{table:loss}, excluding $\mathcal{L}_\text{EA}$ leads to constrained performance across various metrics, indicating the importance of perceptual guidance. Conversely, omitting $\mathcal{L}_\mathcal{G}$ yields superior performance on perceptual quality metrics but suboptimal results on other criteria. The integration of all loss components achieves optimal performance with substantial improvements across multiple metrics.

\vspace{-2mm}
\section{Conclusion and Limitation}

We propose SODiff, a novel one-step diffusion model that extracts semantically rich features with generative priors from over-compressed images to effectively guide the restoration process. It identifies compression severity and selects appropriate timesteps, recovering missing structures and blurred details under extreme compression conditions. 

However, when confronting severe chroma subsampling in extreme cases, SODiff remains susceptible to color shifts despite recovering textural details. This issue will be discussed in the supplementary material. Additionally, separate training of SAIPE (trained from scratch) and diffusion components (fine-tuned from pre-trained weights) is required due to convergence instability, reducing training efficiency.

\bigskip
% \noindent Thank you for reading these instructions carefully. We look forward to receiving your electronic files!

\bibliography{main}
\end{document}